\documentclass{article} % For LaTeX2e
\usepackage{iclr2016_workshop,times}
\usepackage[colorlinks=True]{hyperref}
\usepackage{url}
\usepackage{booktabs}
\usepackage{float}
\usepackage{array,multirow,graphicx}
\usepackage[T1]{fontenc}
\usepackage[utf8]{inputenc}
\usepackage{subcaption}

\title{Learning to SMILE(S)}
\author{Stanisław Jastrzębski, Damian Leśniak \& Wojciech Marian Czarnecki 
 \\
Faculty of Mathematics and Computer Science\\
Jagiellonian University \\
Kraków, Poland \\
\texttt{stanislaw.jastrzebski@uj.edu.pl} \\
}

\begin{document}

\maketitle

\begin{abstract}

This paper shows how one can directly apply natural language processing (NLP) methods to classification problems in cheminformatics. 
Connection between these seemingly separate fields is shown by considering standard textual representation of compound, SMILES. 
The problem of activity prediction against a target protein is considered, which is a crucial part of computer aided drug design process. 
Conducted experiments show that this way one can not only outrank state of the art results of hand crafted representations but also
gets direct structural insights into the way decisions are made.

\end{abstract}

\section{Introduction}

Computer aided drug design has become a very popular
technique for speeding up the process of finding new
biologically active compounds by drastically reducing number of compounds to be tested in laboratory.
Crucial part of this process is virtual screening, where one considers a set of molecules and predicts whether 
the molecules will bind to a given protein. This research focuses on ligand-based virtual screening, where the problem
is modelled as a supervised, binary classification task using only knowledge about ligands (drug candidates) rather than using information about the target (protein).

One of the most underrepresented application areas of deep learning (DL) is believed to be cheminformatics~\citep{unterthiner2014deep, bengio2012survey}, 
mostly due the fact that data is naturally represented as graphs and there are little
direct ways of applying DL in such setting~\citep{DBLP:journals/corr/HenaffBL15}. 
%There has been little work applying general deep learning architectures to cheminformatics (especially virtual screening). 
Notable examples of DL successes in this domain
are winning entry to Merck competition in 2012~\citep{DBLP:journals/corr/DahlJS14} and Convolutional Neural Network (CNN) used for improving data representation~\citep{DBLP:journals/corr/DuvenaudMAGHAA15}. 
To the authors best knowledge all of the above methods use hand crafted representations (called fingerprints)
or use DL methods in a limited fashion.
The main contribution of the paper is showing that one can directly apply DL methods (\emph{without} any customization) to the textual representation of compound (where characters are atoms and bonds). This is analogous
to recent work showing that state of the art performance in language modelling can be achieved considering character-level representation of text~\citep{DBLP:journals/corr/KimJSR15, language_model}.

\subsection{Representing molecules}

Standard way of representing compound in any chemical database is called SMILES, which is just a string of atoms and bonds constructing the molecule (see Fig.~\ref{fig:smiles}) using a specific walk over the graph.
Quite surprisingly, this representation is rarely used as a base of machine learning (ML) methods~\citep{Worachartcheewan2014, PMID:20570021}. 

Most of the classical ML models used in cheminformatics (such as Support Vector Machines or Random Forest) work with constant size vector representation through some predefined embedding (called \emph{fingerprints}). 
As a result many such fingerprints have been proposed across the years~\citep{fingerprint1, fingerprint2}. One of the most common ones are the substructural ones - analogous of bag of word representation
in NLP, where fingerprint is defined as a set of graph \emph{templates} (SMARTS), which are then matched against the molecule to produce binary (set of words)
or count (bag of words) representation. One could ask if this is really necessary, having at one's disposal DL methods of feature learning. 

% \begin{figure}[t]
% \begin{minipage}[t]{0.48\textwidth}
% \begin{center}
% \includegraphics[height=0.06\textheight]{fig/hist_dist_from_diam_fix4.png}
% \caption{Histogram of distance from the diameter.} 
% \label{fig:dist}
% \end{center}
% \end{minipage}
% \hfill%
% \begin{minipage}[t]{0.48\textwidth}
% \begin{center}
% \includegraphics[height=0.06\textheight]{fig/smiles.png}
% \end{center}
% \caption[scale=0.43]{SMILES representation of 3-cyanoanisole is \texttt{COc(c1)cccc1C\#N}. Source: wikipedia}
% \label{fig:smiles}
% \end{minipage}
% \hfill%
% 
% \label{fig:B}
% \end{figure}

\subsection{Analogy to sentiment analysis}

\begin{figure}[h]
\begin{minipage}[t]{0.31\textwidth}
\begin{center}
\includegraphics[height=0.07\textheight]{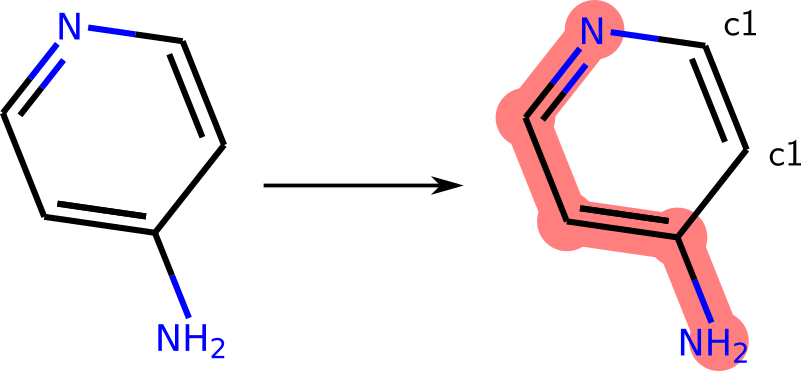}
\end{center}
\caption[scale=0.43]{SMILES produced for the compound in the figure is \texttt{N(c1)ccc1N}.}
\label{fig:smiles}
\end{minipage}
\hfill%
\begin{minipage}[t]{0.31\textwidth}
\begin{center}
\includegraphics[height=0.11\textheight]{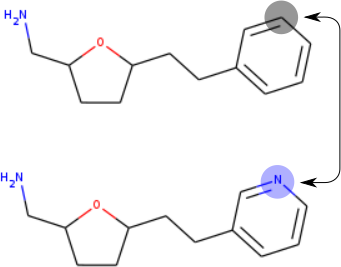}
\caption[scale=0.43]{Substituting highlighted carbon atom with nitrogen renders compound inactive.} 
\label{fig:SERT_pivot}
\end{center}
\end{minipage}
\hfill%
\begin{minipage}[t]{0.31\textwidth}
\begin{center}
\includegraphics[height=0.11\textheight]{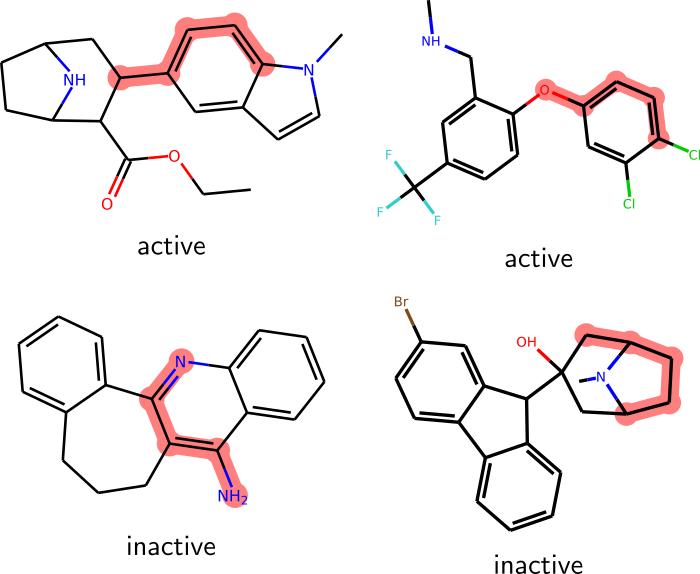}
\end{center}
\caption[scale=0.43]{Visualization of CNN filters of size 5 for active (top row) and inactives molecules.} 
\label{fig:cnn}
\end{minipage}
\label{fig:B}
\end{figure}

The main contribution of this paper is identifying analogy to NLP and specifically sentiment analysis, which is tested by applying state of the art methods~\citep{mesnil} directly to SMILES representation.
The analogy is motivated by two facts. First, small local changes to structure can imply large overall activity change (see Fig.~\ref{fig:SERT_pivot}), just like sentiment is a function of sentiments of different clauses and their connections, which is the main argument for effectiveness of DL methods in this task~\citep{socher}.  %By means of small modifications to their outer cells bacteria can become immune to drugs {\color{red}[\citep{chem-bacteria}]}
Second, perhaps surprisingly, compound graph is almost always nearly a tree. To confirm this claim we calculate molecules diameters, defined as a maximum over all atoms of minimum distance between given atom and the longest carbon chain in the molecule. It appears that in practise analyzed molecules have diameter between 1 and 6 with mean 4. Similarly, despite the way people write down text, human thoughts are not linear, and sentences can have complex clauses. Concluding, in organic chemistry one can make an analogy between longest carbon chain and sentence, where branches stemming out of the longest chain are treated as clauses in NLP.

\section{Experiments}

%\subsection{Data}

Five datasets are considered. Except SMILES, two baseline fingerprint compound representations are used, namely MACCS~\cite{maccs} and Klekota--Roth~\cite{klekota} (KR; considered state of the art of substructural representation~\citep{czarnecki2015bayes}). %This could be desirable to consider more fingerprints (also of different types, not only substructural) in subsequent work.
Each dataset is fairly small (mean size is 3000) and most of the datasets are slightly imbalanced (with mean class ratio around 1:2). It is worth noting that chemical databases are usually fairly big (ChEMBL size is 1.5M compounds), which hints at possible gains by using semi-supervised learning techniques. %Simple preprocessing to SMILES was applied: extraction of n-grams and tokenization ([Na+] becomes a single token). 

Tested models include both traditional classifiers: Support Vector Machine (SVM) using Jaccard kernel,  Naive Bayes (NB), Random Forest (RF) as well as neural network models:
Recurrent Neural Network Language Model~\citep{rnnlm} (RNNLM), Recurrent Neural Network (RNN) many to one classifier, Convolutional Neural Network (CNN) and Feed Forward Neural Network with ReLU activation. Models were selected to fit two criteria: span state of the art models in single target virtual screening~\citep{czarnecki2015bayes, smusz2013multidimensional} and also cover state of the art models in sentiment analysis. %For the second one we follow models selected by \citet{Mesnli}. 
For CNN and RNN a form of data augmentation is used, where for each molecule random SMILES walks are computed and predictions are averaged (not doing so degrades strongly performance, mostly due to overfitting). For methods which are not designed to work on string representation (such as SVM, NB, RF, etc.) SMILES are embedded as n-gram models with simple tokenization (\texttt{[Na+]} becomes a single token). For all the remaining ones, SMILES are treated as strings composed of 2-chars symbols  (thus capturing atom and its relation to the next one).
 
Using RNNLM, $p(\mbox{compound}|\mbox{active})$ and $p(\mbox{compound}|\mbox{inactive})$ are modelled separately and classification is done through logistic regression fitted on top. 
%While this model is not natural for classification we consider it as this is a fully generative model. 
%It would be interesting to investigate more deeply generative models for chemical compounds. Publicly available optimized implemention, faster-rnnlm, is used during experiments [\citep{faster-rnnlm}].
For CNN, purely supervised version of \textsc{context}, current state of the art in sentiment analysis~\citep{DBLP:conf/naacl/Johnson015}, is used. 
Notable feature of the model is working directly on one-hot representation of the data. 
%We leave for future work considering semi-supervised version of \textsc{ConText}.
Each model is evaluated using 5-fold stratified cross validation. Internal 5-fold grid is used for fitting hyperparameters (truncated in the case of deep models).
We use log loss as an evaluation metric to include both classification results as well as uncertainty measure provided by models. Similar conclusions are true for accuracy.
 
 \subsection{Results}
 
 Results are presented in Table~\ref{tab:results}. %The main hypothesis of this paper is confirmed experimentally. 
 First, simple n-gram models (SVM, RF) performance is
 close to hand crafted state of the art representation, which suggests that potentially \emph{any} NLP classifier working on n-gram representation might be applicable. Maybe even more interestingly,
 current state of the art model for sentiment analysis - CNN - despite small dataset size, outperforms (however by a small margin) traditional models. 

\begin{table}[]
\caption{Log-loss ($\pm$ std) of each model for a given protein and representation.}
\label{results-table}
\begin{center}
\scriptsize
\begin{tabular}{lllllll}
\toprule
        & model &           5-HT$_\mathrm{1A}$ &           5-HT$_\mathrm{2A}$ &            5-HT$_7$ &               H1 &             SERT \\
\midrule
	  \parbox[t]{2mm}{\multirow{4}{*}{\rotatebox[origin=c]{90}{SMILES}}} 
           & CNN &  $\mathbf{0.249\pm0.015}$ &  $\mathbf{0.284\pm0.026}$ &  $\mathbf{0.289\pm0.041}$ &  $\mathbf{0.182\pm0.030}$ &  $\mathbf{0.221\pm0.032}$ \\
	 &   SVM &  $0.255\pm0.009$ &  $0.309\pm0.027$ &  $0.302\pm0.033$ &  $0.202\pm0.037$ &  $0.226\pm0.015$ \\
         &   GRU  &  $0.274\pm0.016$ &  $0.340\pm0.035$ &  $0.347\pm0.045$ &  $0.222\pm0.042$ &  $0.269\pm0.032$ \\
	 &  RNNLM &  $0.363\pm0.020$ &  $0.431\pm0.025$ &  $0.486\pm0.065$ &  $0.283\pm0.066$ &  $0.346\pm0.102$ \\
	  \midrule
	\parbox[t]{2mm}{\multirow{4}{*}{\rotatebox[origin=c]{90}{KRFP}}}
    & SVM &  $0.262\pm0.016$ &  $0.311\pm0.021$ &  $0.326\pm0.035$ &  $0.188\pm0.022$ &  $0.226\pm0.014$ \\
   &   RF &  $0.264\pm0.029$ &  $0.297\pm0.012$ &  $0.322\pm0.038$ &  $0.210\pm0.015$ &  $0.228\pm0.022$ \\
   &   NN &  $0.285\pm0.026$ &  $0.331\pm0.015$ &  $0.375\pm0.072$ &  $0.232\pm0.034$ &  $0.240\pm0.024$ \\
   &   NB  &  $0.634\pm0.045$ &  $0.788\pm0.073$ &  $1.201\pm0.315$ &  $0.986\pm0.331$ &  $0.726\pm0.066$ \\
      \midrule
    \parbox[t]{2mm}{\multirow{4}{*}{\rotatebox[origin=c]{90}{MACCS}}}
   &   SVM &  $0.310\pm0.012$ &  $0.339\pm0.017$ &  $0.382\pm0.019$ &  $0.237\pm0.027$ &  $0.280\pm0.030$ \\
   &  RF  &  $0.261\pm0.008$ &  $0.294\pm0.015$ &  $0.335\pm0.034$ &  $0.202\pm0.004$ &  $0.237\pm0.029$ \\
   &  NN  &  $0.377\pm0.005$ &  $0.422\pm0.025$ &  $0.463\pm0.047$ &  $0.278\pm0.027$ &  $0.369\pm0.020$ \\
   &  NB  &  $0.542\pm0.043$ &  $0.565\pm0.014$ &  $0.660\pm0.050$ &  $0.477\pm0.042$ &  $0.575\pm0.017$ \\
\bottomrule
\end{tabular}
\end{center}
\label{tab:results}
\end{table}
 
%We have included in the table only best n-gram range (3, 7) and best model (SVM with Jaccard kernel) combination as our baseline. This model was selected from grid of 10 different
%n-gram ranges and 4 models (SVM, Neural Network, Random Forest and Naive Bayes) via pairwise win ranking.
 
Hyperparameters selected for CNN (\textsc{context}) are similar to the parameters reported in~\citep{DBLP:conf/naacl/Johnson015}. Especially
the maximum pooling (as opposed to average pooling) and moderately sized regions (5 and 3) performed best (see Fig. \ref{fig:cnn}).
This effect for NLP is strongly correlated with the fact that small portion of sentence can contribute strongly to overall sentiment, thus confirming claimed molecule-sentiment analogy.

RNN classifier's low performance can be attributed to small dataset sizes, as commonly RNN are applied to significantly larger volumes of data~\citep{strategies}. One alternative is to consider semi-supervised version of RNN~\citep{NIPS2015_5949}. Another problem is that compound activity prediction requires remembering very long interactions, especially that neighbouring atoms in SMILES walk are often disconnected in the original molecule.

\section{Conclusions}

This work focuses on the problem of compounds activity prediction without hand crafted features used to represent complex molecules.
Presented analogies with NLP problems, and in particular sentiment analysis, followed by experiments performed with the use of state of the art
methods from both NLP and cheminformatics seem to confirm that one can actually learn directly from raw string representation of SMILES instead
of currently used embedding. In particular, performed experiments show that despite being trained on relatively small datasets, 
CNN based solution can actually	 outperform state of the art methods based on structural fingerprints in ligand-based virtual screening task.
At the same time it gives possibility to easily incorporate unsupervised and semi-supervised techniques into the models, making use of huge databases of chemical compounds.
It appears, that cheminformatics can strongly benefit from NLP and further research in this direction should be conducted.

\section*{Acknowledgments}

First author was supported by Grant No.~DI 2014/016644 from Ministry of Science and Higher Education, Poland.

%Conducted experiments state of the art performance can be achieved using NLP methods working directly on string representation of the compound, rather than
%using hand crafted features. We have shown strong analogy between sentiment classification and activity prediction problem. Most promising avenue to pursue next is
%trying semi-supervised learning (using RNN or CNN models), as there is a drastic disproportion between labeled and unlabeled data, again drawing parallel between NLP and cheminformatics research fields.

\bibliography{paper}	
\bibliographystyle{paper}

\end{document}